\documentclass[10pt,twocolumn,letterpaper]{article}
\usepackage[accsupp]{axessibility}

\usepackage{cvpr}      

\usepackage{graphicx}
\usepackage{amsmath}
\usepackage{arydshln}
\usepackage{bbm}
\usepackage{amssymb}
\usepackage{booktabs}
\usepackage{multicol}
\usepackage{multirow}
\usepackage{colortbl}
\usepackage{xcolor}         
\usepackage{color}
\definecolor{mygray}{gray}{.9}
\definecolor{mycyan}{cmyk}{.3,0,0,0}
\definecolor{light-gray}{gray}{0.5}
\definecolor{c0}{RGB}{146,208,80}
\definecolor{c1}{RGB}{0,176,80}
\definecolor{c2}{RGB}{102,153,0} 
\definecolor{c3}{RGB}{0,0,255}
\definecolor{ACM-purple}{RGB}{121,29,125}

\newcommand{\tabincell}[2]{\begin{tabular}{@{}#1@{}}#2\end{tabular}}
\usepackage[pagebackref,breaklinks,colorlinks]{hyperref}
\usepackage[accsupp]{axessibility}

\usepackage[capitalize]{cleveref}
\crefname{section}{Sec.}{Secs.}
\Crefname{section}{Section}{Sections}
\Crefname{table}{Table}{Tables}
\crefname{table}{Tab.}{Tabs.}

\def\cvprPaperID{5223} 
\def\confName{CVPR}
\def\confYear{2022}

\begin{document}

\title{Wavelet Knowledge Distillation: Towards Efficient Image-to-Image Translation}

\author{Linfeng Zhang$^1$~~~~Xin Chen$^2$~~~~Xiaobing Tu$^3$~~~~Pengfei Wan$^3$~~~~Ning Xu$^3$~~~~ Kaisheng Ma$^1$\thanks{~Corresponding author.}\\
Tsinghua Univeristy$^1$~~ Intel Corporation$^2$~~ Kuaishou Technology$^3$\thanks{~This project is funded by Kuaishou Research Program. This work was done while Linfeng Zhang was an intern at Y-tech KuaiShou.}\\
{\tt\small zhang-lf19@mails.tsinghua.edu.cn,xin.chen@intel.com,\{tuxiaobing,wanpengfei\}@kuaishou.com
}\\
{\tt\small ningxu01@gmail.com, kaisheng@mail.tsinghua.edu.cn}
}
\maketitle


\begin{abstract} 
    Remarkable achievements have been attained with Generative Adversarial Networks (GANs) in image-to-image translation. 
    However, due to a tremendous amount of parameters, state-of-the-art GANs usually suffer from low efficiency and bulky memory usage.
    To tackle this challenge, firstly, this paper investigates GANs performance from a frequency perspective. The results show that GANs, especially small GANs lack the ability to generate high-quality high frequency information.
    To address this problem, we propose a novel knowledge distillation method referred to as wavelet knowledge distillation.
    Instead of directly distilling the generated images of teachers, wavelet knowledge distillation first decomposes the images into different frequency bands with discrete wavelet transformation and then only distills the high frequency  bands. As a result, the student GAN can pay more attention to its learning on high frequency bands.
    Experiments demonstrate that our method leads to 7.08$\times$ compression and 6.80$\times$ acceleration on CycleGAN with almost no performance drop. 
    Additionally, we have studied the relation between discriminators and generators which shows that the compression of discriminators can promote the performance of compressed generators.
    
\end{abstract}
\section{Introduction}
Tremendous progress has been achieved with Generative adversarial networks~(GANs) in generating high-fidelity, high-resolution, and photo-realistic images and videos with both paired and unpaired datasets~\cite{singan,bigGAN,gan,pix2pix,cyclegan,DBLP:conf/cvpr/LiZ0CHMHWJ21,DBLP:conf/cvpr/JeongKLS21,DBLP:conf/cvpr/RichardsonAPNAS21}. 
The excellent performance of GANs has promoted its application in various image-to-image translation tasks, such as image style transfer~\cite{stylegan,styleganv2} and super-resolution~\cite{srgan}. 
Compared with other tasks such as image classification and object detection, image-to-image generation is more complex since it has a much larger output space.  
As a consequence, existing GANs always have 
high computational demands and a huge amount of parameters,
which lead to inefficient inference and intolerant memory footprint, and limit their usage in resource-constrained platforms.

Knowledge distillation~(KD)
has been an effective tool for improving the performance of small models \cite{model_compression,distill_hinton}.
By imitating the prediction results and the intermediate features from a cumbersome teacher model,  the performance of a lightweight student model can be improved significantly.
Following previous knowledge distillation methods on classification~\cite{kd_crd}, object detection~\cite{detectiondistillation}, semantic segmentation~\cite{structured_kd} and action recognition~\cite{liu2019paying}, some recent research has tried to directly apply knowledge distillation to GANs. Unfortunately, most of them obtain very limited and even negative effects~\cite{gan_compress,spkd_gan}. 

\begin{figure}
    \begin{center}
        \includegraphics[width=7.5cm]{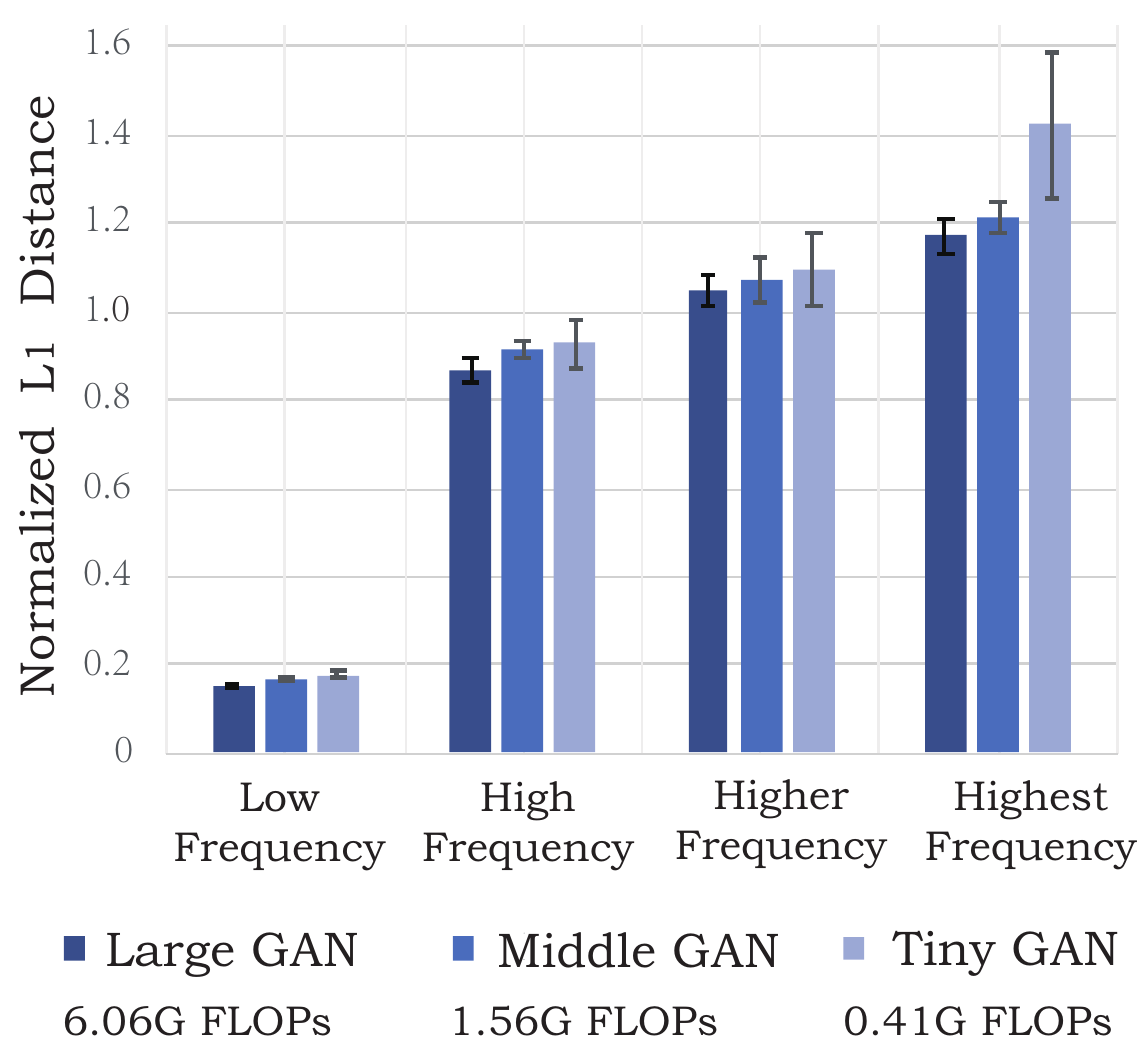}
    \end{center}
    \vspace{-0.40cm}
    \caption{\label{fig1} Normalized $L_1$ distance between the images generated by GANs and the ground truth images on different frequency bands. Different colors indicate GANs with different FLOPs. Results are averaged over 8 trials on Edge$\rightarrow$Shoe dataset. 
    \vspace{-0.25cm}
    }

\end{figure}
\begin{figure*}
    \begin{center}
        \includegraphics[width=18cm]{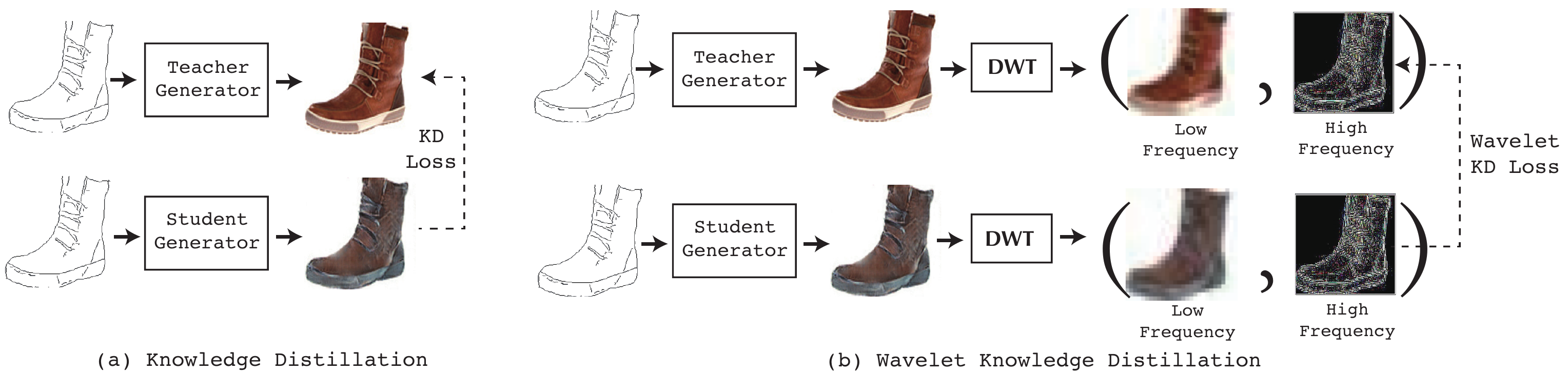}
        \vspace{-0.7cm}
    \end{center}
    \caption{\label{fig:compare} Comparison between knowledge distillation~\cite{distill_hinton} (subfigure a) and the proposed wavelet knowledge distillation (subfigure b) on Edges$\rightarrow
    $Shoes. 
    Wavelet knowledge distillation first applies discrete wavelet transformation (DWT) to the generated images and then only minimizes the difference on high frequency bands. }
    \vspace{-0.2cm}
\end{figure*}

\emph{Why does KD not work well on GAN?} In this paper, we first study this question from a frequency perspective with the following experiment. Firstly, discrete wavelet transformation (DWT) is utilized to decompose the generated images and the ground truth images into different frequency bands. Then, we compute the normalized $L_1$-norm distance on each frequency band respectively\footnote{Details of this experiment can be found in the supplementary material.}. As shown in Figure~\ref{fig1}, all the GANs achieve very low error on the low frequency band but fail in the generation on high frequency bands, which is in line with the observation that images generated by GANs do not have good details.
Besides, it is observed that compared with the large GAN, the tiny GAN achieves comparable performance on the low frequency band but much worse performance on high frequency bands. These two observations demonstrate that more attention should be paid to the high frequency during GAN compression.

However, naive knowledge distillation in GANs application directly minimizes the difference between the images generated by students and teachers and ignores the priority of high frequency.
Motivated by these observations, we propose wavelet knowledge distillation, which highlights students learning on the high frequency in knowledge distillation. As shown in Figure~\ref{fig:compare}, we first apply discrete wavelet transformation to decompose the images generated by teachers and students into different frequency bands and then only minimize the $L_1$ loss on the high frequency bands. 

Abundant experiments on both paired and unpaired image-to-image translation demonstrate the effectiveness of our method both quantitatively and qualitatively.
On Horse$\rightarrow$Zebra and Zebra$\rightarrow$Horse datasets, our method leads to 7.08$\times$ compression and 6.80$\times$ acceleration on CycleGAN with almost no performance drop. In the discussion section, we have further studied the effectiveness of different frequency bands and the influence of knowledge distillation schemes.
Additionally, studies on the relation between discriminators and generators in model compression have also been introduced, showing that the compression of discriminators can significantly promote the performance of compressed generators. 

Our main contributions can be summarized as follows.
\begin{itemize}
    \item We have analyzed the performance of GANs from a frequency perspective, which quantitatively shows that GAN, especially small GAN lacks the ability to generate high-quality high frequency information in images. 
    
    \item Based on the above observation, wavelet knowledge distillation is proposed to address this issue by only distilling the high frequency information, instead of all the information from images generated by the teacher.
    
    \item Quantitative and qualitative results on three models and eight datasets with six comparison  methods have demonstrated the effectiveness of our method.
    
    \item We have studied the relation between discriminators and generators during model compression. It shows that compression on discriminators is necessary for maintaining its competition with compressed generators in adversarial learning. which further benefits the performance of generators. 
    
\end{itemize}

\section{Related Work}
\subsection{Image-to-Image Translation}
Generative adversarial networks have shown powerful ability in formulating and generating high-fidelity, high-resolution, and photo-realistic images and videos, and thus become the dominant models in image-to-image translation~\cite{gan,videogan,new_style1,new_style2,new_style3,new_style4,new_style5}.
Pix2Pix is first proposed to apply conditional generative adversarial networks for paired image-to-image translation~\cite{pix2pix}. Then, Pix2PixHD is proposed to generate images with higher resolution with coarse to fine generators and multi-scale discriminators~\cite{pix2pixHD}. A more challenging task is to perform image-to-image translation with unpaired datasets. CycleGAN tackles this challenge by introducing the cycle-consistency loss, which reconstructs the generated image to the input domain~\cite{cyclegan}. Then, attention CycleGAN is proposed to find the crucial pixels in the images with an attention module~\cite{attentiongan}. Spade module is introduced in GauGAN to avoid the loss of semantic information in batch normalization layers~\cite{gaugan}.
Recently, researchers find that the perfect reconstruction in CycleGAN may be too difficult to achieve~\cite{breakingcycle}. To address this issue, Park~\emph{et al.} introduce the patch-wise contrastive learning which improves the generation quality, stabilizes the training process, and reduces the training time, simulanteously~\cite{park2020contrastive}.
The high-resolution and photo-realistic generated images come at the expense of intensive computation and massive parameters. To tackle this challenge, fruitful compression methods such as network pruning and network architecture search have been proposed recently. Li~\emph{et al.} propose the GAN Compression, which applies once-for-all search to find the best tiny GAN architecture~\cite{gan_compress}. Jin~\emph{et~al.} introduce an inception-based residual block into generators and further compress them with channel pruning~\cite{teacher_do_more_gankd}. Liu~\emph{et al.} propose the Content-Aware GAN Compression, which enables GANs to maintain the content of crucial regions during compression~\cite{liu2021content}. Li~\emph{et al.} propose  to revisit the role of discriminator in GAN compression with a selective activation discriminator~\cite{revisit_discriminator}.

\subsection{Knowledge Distillation}
Knowledge distillation, which aims to facilitate the training of a lightweight student model under the supervision of a cumbersome teacher model, has been considered as an effective approach for both model compression and model accuracy boosting. The idea of employing a teacher model to train a student model is first proposed by Buciluǎ~\emph{et~al.} for ensemble model compression~\cite{model_compression}. Then Hinton~\emph{et al.} propose the concept of knowledge distillation, which introduces a hyper-parameter named temperature in softmax to soften the distribution of teacher logits~\cite{distill_hinton}.
Recently, abundant methods have been proposed to distill the knowledge in the intermediate features~\cite{tofd,attentiondistillation} and their relation~\cite{relational_kd, relational_kd2}.
Besides image classification, recent works have also successfully applied knowledge distillation to more challenging tasks such as object detection~\cite{kd_detection4,detectiondistillation}, semantic segmentation~\cite{structured_kd}, pre-trained language models~\cite{kd_bert1,kd_bert2}, machine translation~\cite{lin2020weight}, distributed training~\cite{distillationdefense}, multi-exit models~\cite{Zhang2019SCANAS} and so on.

However, the effect of knowledge distillation on GANs for image-to-image translation has not been well-studied. Existing research shows that directly minimizing the distance between the generated images of students and teachers does not improve but sometimes harms the performance of students~\cite{spkd_gan}. A few previous methods have tried to apply the classification-based knowledge distillation to image-to-image translation but earned very limited improvements.  
For instance, Li~\emph{et al.} propose to minimize the distance between intermediate features of teacher and student GANs~\cite{gan_compress} and  Li~\emph{et al.} have tried to distill the semantic relation between the features of different patches~\cite{spkd_gan}. Recently, Chen~\etal propose an overall knowledge distillation framework on GANs by distilling both the generator and the discriminator~\cite{distill_portable_gan}. Jin~\etal 
~propose to distill generators with global kernel alignment on intermediate features~\cite{teacher_do_more_gankd}, which boosts student performance without introducing additional layers.
The main difference between our method and the previous GAN knowledge distillation methods is that our method distills the generated images
instead of the intermediate features. As a result, our method is orthogonal to the previous methods, and it can be utilized with previous methods to achieve better performance.

\subsection{Wavelet Analysis in Deep Learning}
Compared with the other frequency analysis methods such as Fourier analysis, wavelet transformation can capture both the spatial and the frequency information in the sign and is thus considered as a more effective method in image processing~\cite{wavelet}.
Along with the success of deep learning, fruitful methods have been proposed to apply wavelet methods into neural networks for different targets. 
Williams~\etal propose the wavelet pooling which replaces the max and average pooling with discrete wavelet transformation to preserve the global information of images during down sampling~\cite{wavepool}. 
Chen~\emph{et al.} propose wavelet-like auto-encoder, which compresses the original image into two low-resolution images to accelerate the inference computation~\cite{wae}.  Liu~\emph{et al.} introduce wavelet transformation to the convolutional neural networks to leverage the spectral information in texture classification~\cite{wavelet_texture}.

Recent works have also applied wavelet analysis to the image-to-image translation tasks. Huang~\emph{et al.} first propose the wavelet-SRNet, which performs single image super-resolution by predicting the wavelet coefficients of high-resolution images~\cite{huang2017wavelet}.
Inspired by the architecture of U-Net, Liu~\emph{et al.} apply the wavelet package in convolutional neural networks to obtain a large receptive field efficiently~\cite{liu2018multi}. 
To the best of our knowledge, this paper is the first work which applies wavelet analysis to knowledge distillation and GANs compression.

\begin{table*}
\footnotesize
\caption{\label{tab:quan1} Experiment results on paired image-to-image translation on Edges$\rightarrow$Shoes with Pix2Pix and Pix2PixHD. A lower FID is better performance. 
 $\Delta$ indicates the performance improvements compared with the origin student. Each result is averaged over 8 trials.
 \vspace{-0.5cm}
}

\begin{center}
  \setlength{\tabcolsep}{1.2mm}{ \begin{tabular}{c c l c c l c c l c c l}

   \multicolumn{5}{c}{\texttt{Pix2PixHD}}&&
   \multicolumn{5}{c}{\texttt{Pix2Pix}}\\
    \toprule
    \multirow{2}{*}{\#Params~(M)}& \multirow{2}{*}{FLOPs~(G)}&\multirow{2}{*}{Method} & \multicolumn{2}{c}{Metric}& &\multirow{2}{*}{\#Params~(M)}& \multirow{2}{*}{FLOPs~(G)}&\multirow{2}{*}{Method} & \multicolumn{2}{c}{Metric}\\
   \cmidrule{4-5}\cmidrule{10-11}
   &&&FID$\downarrow$&$\Delta \uparrow$&&   &&&FID$\downarrow$&$\Delta \uparrow$\\
   \cmidrule{1-5} \cmidrule{7-11}

45.59&48.36&Teacher&41.59$\pm$0.42&--&&54.41&6.06&Teacher&59.70$_{\pm0.91}$&--\\       \cmidrule{1-5} \cmidrule{7-11}
    \multirow{6}{*}{\tabincell{c}{1.61\\ $_{28.32\times}$}}&\multirow{6}{*}{ \tabincell{c}{1.89\\$_{25.59\times}$}}&Origin Student&44.64$_{\pm0.54}$&--&&    
    \multirow{6}{*}{\tabincell{c}{13.61\\ $_{4.00\times}$}}&\multirow{6}{*}{\tabincell{c}{1.56 \\$_{3.88\times}$}}&Origin Student&85.06$_{\pm0.98}$&--\\

    &&Hinton~\emph{et al.}~\cite{distill_hinton}&45.31$_{\pm0.63}$&-0.67
    &&&&Hinton~\emph{et al.}~\cite{distill_hinton}&86.97$_{\pm3.49}$&-1.91\\
    
    &&Zagoruyko~\emph{et al.}~\cite{attentiondistillation}&44.21$_{\pm0.72}$&0.43
    &&&&Zagoruyko~\emph{et al.}~\cite{attentiondistillation}&84.25$_{\pm2.08}$&0.81\\
    
    &&Li and Lin~\emph{et al.}~\cite{gan_compress}&44.03$_{\pm0.41}$&0.61
    &&&&Li and Lin~\emph{et al.}~\cite{gan_compress}&83.63$_{\pm3.12}$&1.43\\

    &&Li and Jiang~\emph{et al.}~\cite{spkd_gan}&43.90$_{\pm0.36}$&0.74
    &&&&Li and Jiang~\emph{et al.}~\cite{spkd_gan}&84.01$_{\pm2.31}$&1.05\\

    &&Jin~\emph{et al.}~\cite{teacher_do_more_gankd}&43.97$_{\pm0.17}$&0.67
    &~~~~&&&Jin~\emph{et al.}~\cite{teacher_do_more_gankd}&84.39$_{\pm3.62}$&0.67\\

    &&Ahn~\emph{et al.}~\cite{kd_variational}&44.53$_{\pm0.48}$&0.11
    &&&&Ahn~\emph{et al.}~\cite{kd_variational}&84.92$_{\pm0.78}$&0.14\\
    
    &&\textbf{Ours}\cellcolor{mygray}&\cellcolor{mygray}\textbf{42.53$_{\pm\mathbf{0.29}}$}&\cellcolor{mygray}\textbf{2.11}
    &&&&\textbf{Ours}\cellcolor{mygray}&\cellcolor{mygray}\textbf{80.13$_{\pm\mathbf{2.18}}$}&\cellcolor{mygray}\textbf{4.93}\\

\bottomrule
\end{tabular}}
\end{center}
\end{table*}

\begin{table*}
\footnotesize
\caption{\label{tab:quan2} Experiment results on unpaired image-to-image translation on Horse$\rightarrow$Zebra and Zebra$\rightarrow$Horse with CycleGAN. A lower FID is better. $\Delta$ indicates the performance improvements compared with the origin student.
Each result is averaged over 8 trials.  \vspace{-0.5cm}
}

\begin{center}
  \setlength{\tabcolsep}{1.0mm}{ \begin{tabular}{c c l c c l c c l c c l}

   \multicolumn{5}{c}{\texttt{Horse$\rightarrow$Zebra}}&&
   \multicolumn{5}{c}{\texttt{Zebra$\rightarrow$Horse}}\\
    \toprule
    \multirow{2}{*}{\#Params~(M)}& \multirow{2}{*}{FLOPs~(G)}&\multirow{2}{*}{Method} & \multicolumn{2}{c}{Metric}& &\multirow{2}{*}{\#Params~(M)}& \multirow{2}{*}{FLOPs~(G)}&\multirow{2}{*}{Method} & \multicolumn{2}{c}{Metric}\\
   \cmidrule{4-5}\cmidrule{10-11}
   &&&FID$\downarrow$&$\Delta \uparrow$&&   &&&FID$\downarrow$&$\Delta \uparrow$\\
   \cmidrule{1-5} \cmidrule{7-11}

11.38&49.64&Teacher&61.34$_{\pm4.35}$&--&&11.38&49.64&Teacher&138.07$_{\pm4.01}$&--\\       \cmidrule{1-5} \cmidrule{7-11}

    \multirow{6}{*}{\tabincell{c}{0.72\\ $_{15.81\times}$}}&\multirow{6}{*}{ \tabincell{c}{3.35\\$_{14.82\times}$}}&Origin Student&85.04$_{\pm6.88}$&--&&    
    \multirow{6}{*}{\tabincell{c}{0.72\\ $_{15.81\times}$}}&\multirow{6}{*}{\tabincell{c}{3.35 \\$_{14.82\times}$}}&Origin Student&152.67$_{\pm9.63}$&--\\

    &&Hinton~\emph{et al.}~\cite{distill_hinton}&84.08$_{\pm3.78}$&0.96
    &&&&Hinton~\emph{et al.}~\cite{distill_hinton}&148.64$_{\pm1.62}$&4.03\\
    
    &&Zagoruyko~\emph{et al.}~\cite{attentiondistillation}&81.24$_{\pm2.01}$&3.80
    &&&&Zagoruyko~\emph{et al.}~\cite{attentiondistillation}&148.92$_{\pm1.20}$&3.75\\
    
    &&Li and Lin~\emph{et al.}~\cite{gan_compress}&83.97$_{\pm5.01}$&1.07
    &&&&Li and Lin~\emph{et al.}~\cite{gan_compress}&151.32$_{\pm2.31}$&1.35\\

    &&Li and Jiang~\emph{et al.}~\cite{spkd_gan}&81.74$_{\pm4.65}$&3.30
    &&&&Li and Jiang~\emph{et al.}~\cite{spkd_gan}&151.09$_{\pm3.67}$&1.58\\

    &&Jin~\emph{et al.}~\cite{teacher_do_more_gankd}&82.37$_{\pm8.56}$&2.67
    &~~~~&&&Jin~\emph{et al.}~\cite{teacher_do_more_gankd}&149.73$_{\pm3.94}$&2.94\\

    &&Ahn~\emph{et al.}~\cite{kd_variational}&82.91$_{\pm2.41}$&2.13
    &&&&Ahn~\emph{et al.}~\cite{kd_variational}&150.31$_{\pm3.55}$&2.36\\
    
    &&\textbf{Ours}\cellcolor{mygray}&\cellcolor{mygray}\textbf{77.04$_{\pm\mathbf{3.52}}$}&\cellcolor{mygray}\textbf{8.00}
    &&&&\textbf{Ours}\cellcolor{mygray}&\cellcolor{mygray}\textbf{146.01$_{\pm\mathbf{1.86}}$}&\cellcolor{mygray}\textbf{6.66}\\
    
    &&\textbf{Ours + Li and Lin~\emph{et al.}}\cellcolor{mygray}&\cellcolor{mygray}\textbf{76.40$_{\pm\mathbf{3.17}}$}&\cellcolor{mygray}\textbf{8.64}
    
    &&&&\textbf{Ours + Li and Lin~\emph{et al.}}\cellcolor{mygray}&\cellcolor{mygray}\textbf{145.96$_{\mathbf{\pm1.92}}$}&\cellcolor{mygray}\textbf{6.71}\\
    \midrule

    \multirow{6}{*}{\tabincell{c}{1.61\\ $_{7.08\times}$}}&\multirow{6}{*}{ \tabincell{c}{7.29\\$_{6.80\times}$}}&Origin Student&70.54$_{\pm9.63}$&--&&    
    \multirow{6}{*}{\tabincell{c}{1.61\\ $_{7.08\times}$}}&\multirow{6}{*}{\tabincell{c}{7.29 \\$_{6.80\times}$}}&Origin Student&141.86$_{\pm1.57}$&--\\

    &&Hinton~\emph{et al.}~\cite{distill_hinton}&70.35$_{\pm3.27}$&0.19
    &&&&Hinton~\emph{et al.}~\cite{distill_hinton}&142.03$_{\pm1.61}$&-0.17\\
    
    &&Zagoruyko~\emph{et al.}~\cite{attentiondistillation}&67.51$_{\pm4.57}$&3.03
    &&&&Zagoruyko~\emph{et al.}~\cite{attentiondistillation}&141.23$_{\pm1.88}$&0.63\\
    
    &&Li and Lin~\emph{et al.}~\cite{gan_compress}&68.58$_{\pm4.31}$&1.96
    &&&&Li and Lin~\emph{et al.}~\cite{gan_compress}&141.32$_{\pm1.27}$&0.54\\

    &&Li and Jiang~\emph{et al.}~\cite{spkd_gan}&68.94$_{\pm2.98}$&1.60
    &&&&Li and Jiang~\emph{et al.}~\cite{spkd_gan}&151.09$_{\pm3.67}$&1.58\\

    &&Jin~\emph{et al.}~\cite{teacher_do_more_gankd}&67.31$_{\pm3.01}$&3.23
    &~~~~&&&Jin~\emph{et al.}~\cite{teacher_do_more_gankd}&140.98$_{\pm1.41}$&0.88\\

    &&Ahn~\emph{et al.}~\cite{kd_variational}&69.32$_{\pm5.89}$&1.22
    &&&&Ahn~\emph{et al.}~\cite{kd_variational}&141.50$_{\pm2.51}$&0.36\\

    &&\textbf{Ours}\cellcolor{mygray}&\cellcolor{mygray}\textbf{61.65$_{\pm\mathbf{4.73}}$}&\cellcolor{mygray}\textbf{8.89}
    &&&&\textbf{Ours}\cellcolor{mygray}&\cellcolor{mygray}\textbf{138.84$_{\pm\mathbf{1.47}}$}&\cellcolor{mygray}\textbf{3.02}\\
    
    &&\textbf{Ours + Li and Lin~\emph{et al.}}\cellcolor{mygray}&\cellcolor{mygray}\textbf{60.13$_{\mathbf{\pm4.08}}$}&\cellcolor{mygray}\textbf{10.41}
    
    &&&&\textbf{Ours + Li and Lin~\emph{et al.}}\cellcolor{mygray}&\cellcolor{mygray}\textbf{138.52$_{\mathbf{\pm0.95}}$}&\cellcolor{mygray}\textbf{3.34}\\
\bottomrule
\vspace{-0.5cm}
\end{tabular}}
\end{center}
\end{table*}    

\begin{figure*}
    \begin{center}
        \includegraphics[width=17.5cm]{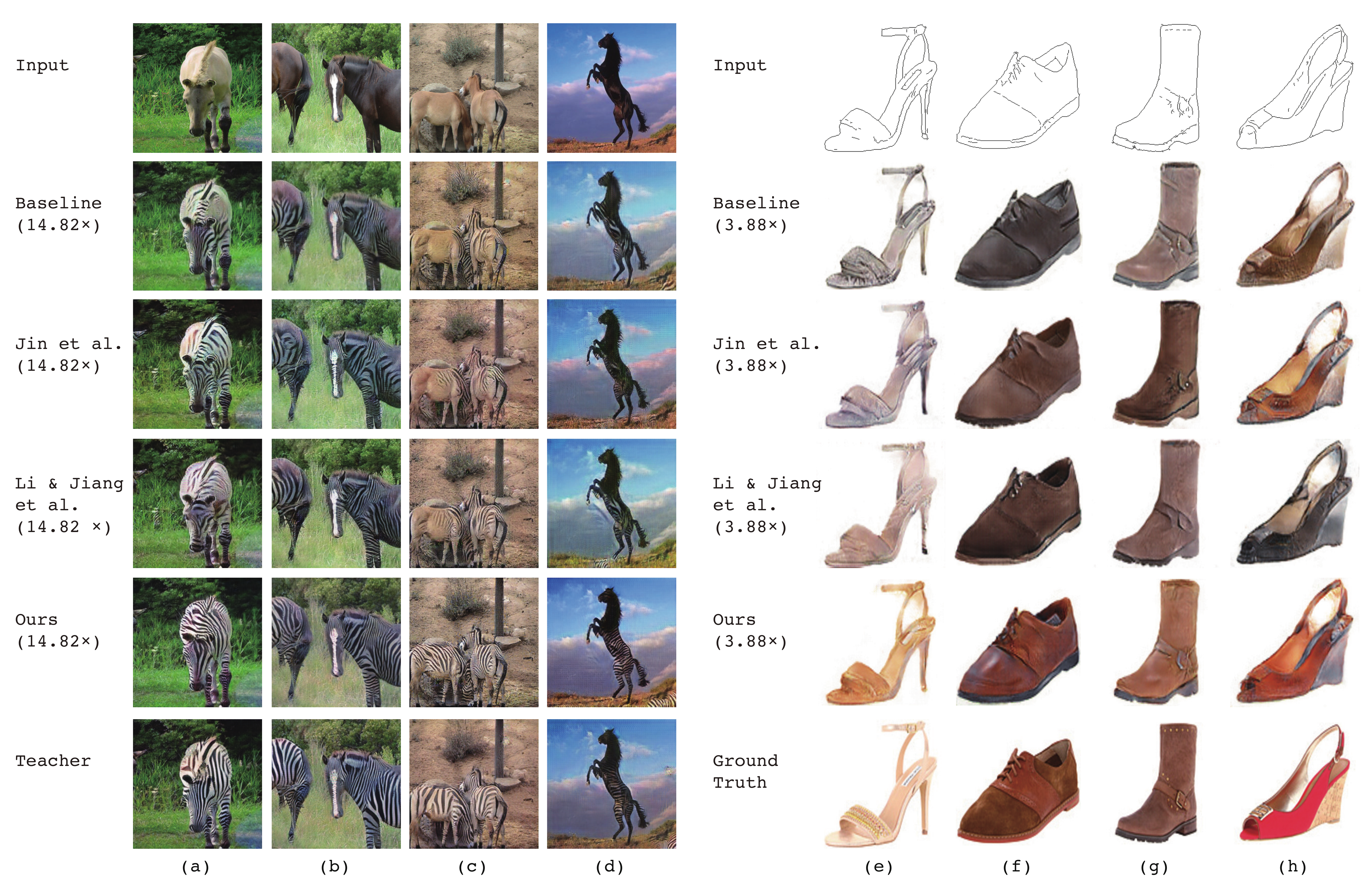}
    \end{center}
    \caption{\label{fig:qualitive} Qualitative results on Horse$\rightarrow$Zebra with CycleGAN (a-d) and Edges$\rightarrow$Shoes with Pix2Pix (e-h). Numbers in the brackets indicate the acceleration ratio compared with their teachers. ``Baseline'' indicates the students trained without knowledge distillation.  }
\end{figure*}

\section{Methodology}
\subsection{Wavelet Analysis}
Given a function $\psi$, let $\mathcal{X}(\psi)$ be the collection of the dilations and shift of $\psi$:
\begin{equation}
\mathcal{X}(\psi)=\{\psi_{jk}=2^{-j/2}\psi(2^{-j}x-k) |~j,k\in Z\},
\end{equation}
where $\psi$ is the orthogonal wavelet if $\mathcal{X}(\psi)$ forms a basis in $\mathcal{L}_2$ spaces.  Discrete wavelet transformation (DWT) is a mathematical tool for pyramidal image decomposition. With DWT, each image can be decomposed into four bands, including {\small LL, LH, HL and HH}, where {\small LL} indicates the low frequency band and the others are high frequency bands. The {\small LL} band can be further decomposed by DWT into {\small LL2, LH2, HL2 HH2} and so on. Denote DWT as $\Psi(\cdot)$, then the high frequency and the low frequency bands of an image $x$ can be written as $\Psi^\text{H}(x)$ and $\Psi^\text{L}(x)$, respectively. 
More specifically, in this paper, we apply 3-level discrete wavelet transformation in all the experiments. $\Psi^\text{L}(x)$ indicates {\small LL3} band. $\Psi^\text{H}(x)=\{${\small HL3,~ LH3,~ HH3,~ HL2,~ LH2,~ HH2,~ HL1,~ LH1,~ HH1}\}.

\subsection{Knowledge Distillation}

\paragraph{Revisit Knowledge Distillation for Classification} At the beginning of this subsection, we revisit the formulation of knowledge distillation on classification
~\cite{distill_hinton}. Given a set of training samples $\mathcal{X}=\{x_1, x_2, ...,x_n\}$ and their labels $\mathcal{Y}=\{y_1, y_2, ..., y_n\}$, denoting the networks of the student and the teacher as $f_s$ and $f_t$, the loss function of the student can be formulated as $\mathcal{L}_{\text{Student}} = \alpha \cdot \mathcal{L}_{\text{CE}}  + (1-\alpha) \cdot \mathcal{L}_{\text{KD}} $, where $\mathcal{L}_{\text{CE}}$ indicates the cross-entropy loss between the prediction $f(x)$ and its label $y$. $\alpha \in (0, 1]$ is a hyper-parameter to balance two loss items, and $\mathcal{L}_{\text{KD}}$ indicates the knowledge distillation loss. 

On classification tasks, $\mathcal{L}_{\text{KD}}$ can be formulated as 
\begin{equation}
\footnotesize
    \mathcal{L}_{\text{KD}} = \frac{1}{n} \sum_i^n \mathcal{KL}\left(\text{softmax}\left(\frac{{}f_t(x_i)}{\tau}\right), \text{softmax}\left(\frac{f_s(x_i)}{\tau}\right)\right),
\end{equation}
where $\mathcal{KL}$ indicates the Kullback-Leibler divergence, which measures the distance between the categorical probability distribution of students and teachers. $\tau$ is the temperature hyper-parameter in softmax function.

\paragraph{Knowledge Distillation for Image-to-Image Translation} On the task of image-to-image translation, since the prediction result $f(x_i)$ is the value of pixels instead of categorical probability distribution, KL divergence can not be utilized to measure the difference between students and teachers.
A~naive alternative is to replace KL divergence with the $L_1$-norm distance between the generated images from students and teachers. Then, we can extend Hinton knowledge distillation for image-to-image translation, whose loss function can be formulated as 
\begin{equation}
\label{gan_kd}
    \mathcal{L}_{\text{KD}} = \frac{1}{n} \sum_i^n \|(f_t(x_i) - f_s(x_i) \|_1.
\end{equation}
Besides Hinton knowledge distillation, there are also abundant feature knowledge distillation methods which can be applied to image-to-image translation directly. Since our method is not feature-based, we do not introduce them here.

\paragraph{Wavelet Knowledge Distillation}
Based on the above notations, now we can introduce the proposed wavelet knowledge distillation, which only minimizes the difference on the high frequency between students and teachers. Its loss function $\mathcal{L}_{\text{WKD}}$ can be formulated as 
\begin{equation}
\label{our loss}
        \mathcal{L}_{\text{WKD}} = \frac{1}{n} \sum_i^n \|(\Psi^H \circ f_t)(x_i) - (\Psi^H \circ f_s)(x_i) \|_1.
\end{equation}
On unpaired image-to-image translation models such as CycleGAN, there are sometimes two generators for the two translation directions. In this circumstance, the proposed wavelet knowledge distillation loss can be applied to the two directions simultaneously. 

The overall training loss can be formulated as $\mathcal{L}_{\text{overall}}=\mathcal{L}_{\text{origin}}+ \alpha \cdot \mathcal{L}_{\text{WKD}}$, where $\mathcal{L}_{\text{origin}}$ indicates the original training loss of different models, such as the adversarial learning loss and recycling loss. $\alpha$ is the hyper-parameter to balance the two loss functions. Hyper-parameter sensitivity studies have been given in the supplementary material.

\begin{figure*}
    \centering
    \includegraphics[width=17.5cm]{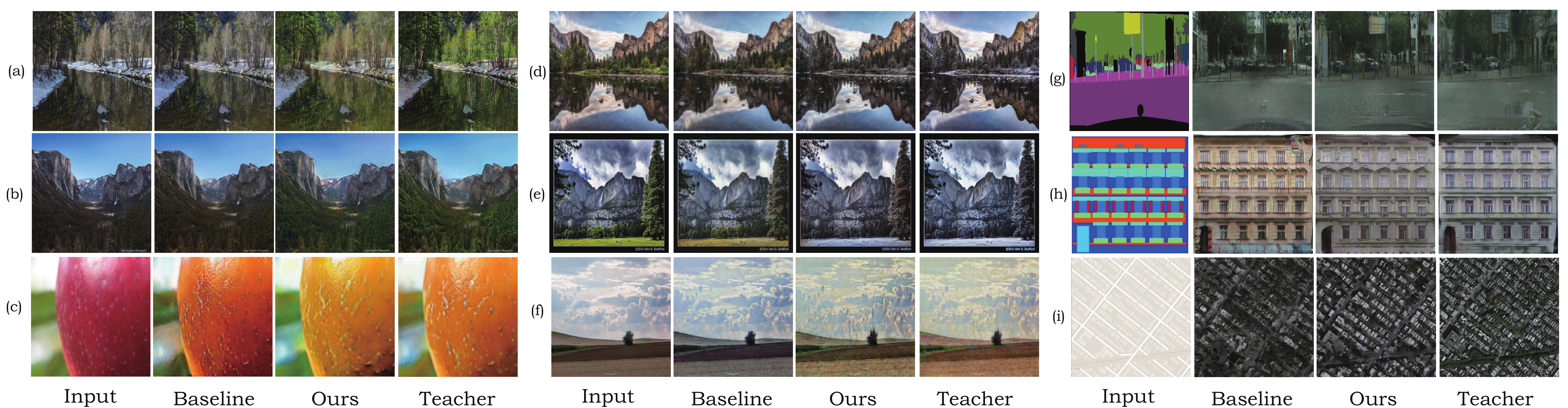}
    \caption{Qualitative experiments on the other datasets: Winter$\rightarrow$Summer (subfig. a-b),  Summer$\rightarrow$Winter (subfig. d-e),  Apple$\rightarrow$Orange (subfig. c), Photo$\rightarrow$Monet (subfig. f), Cityscapes (subfig. g), Facades (subfig. h) and Maps (subfig. i).}
    \label{fig:more}
    \vspace{-0.25cm}
\end{figure*}

\begin{table}
\caption{\label{tab:cityscapes} Paired image-to-image translation experiment results on Cityscapes with Pix2Pix. A higher mIoU is better. $\Delta$ indicates the performance improvements compared with the origin student.
Each experiment is averaged over 8 trials.}
\vspace{-0.3cm}
\begin{center}
\footnotesize
        \setlength{\tabcolsep}{1.0mm}{
   \begin{tabular}{ cc l c c c c }
   
   \toprule \multirow{2}{*}{\#Params~(M)}& \multirow{2}{*}{FLOPs~(G)}&\multirow{2}{*}{Method} & \multicolumn{2}{c}{Metric}\\
    \cmidrule{4-5}
   &&&mIoU$\uparrow$&$\Delta \uparrow$\\
   \midrule
   54.41&96.97&Teacher&46.51$\pm$0.32&--\\
    
    \midrule
    \multirow{6}{*}{13.61 \tabincell{c}{$_{4.00\times}$}}&\multirow{6}{*}{\tabincell{c}{24.90 $_{3.88\times}$}}&Origin Student&41.35$_{\pm0.22}$&--\\
    &&Hinton~\emph{et al.}~\cite{distill_hinton}&40.49$_{\pm0.41}$&-0.86\\
    &&Zagoruyko~\emph{et al.}~\cite{attentiondistillation}&40.17$_{\pm0.36}$&-1.18\\
    &&Li and Lin~\emph{et al.}~\cite{gan_compress}&41.52$_{\pm0.34}$&0.17\\
    &&Li and Jiang~\emph{et al.}~\cite{spkd_gan}&41.77$_{\pm0.30}$&0.42\\
    &&Jin~\emph{et al.}~\cite{teacher_do_more_gankd}&41.29$_{\pm0.51}$&-0.06\\
     &&Ahn~\emph{et al.}~\cite{kd_variational}&41.88$_{\pm0.45}$&0.53\\
    &&\textbf{Ours}\cellcolor{mygray}&\cellcolor{mygray}\textbf{42.93$_{\pm\mathbf{0.25}}$}&\cellcolor{mygray}\textbf{1.58}\\
   \bottomrule
\end{tabular}}
    \vspace{-0.25cm}\end{center}
\end{table}

\section{Experiment}
 
\subsection{Experiment Settings}
 
\paragraph{Models and Datasets} Experiments are mainly conducted with three models including Pix2Pix~\cite{pix2pix}, CycleGAN~\cite{cyclegan} and Pix2PixHD~\cite{pix2pixHD}. Our teacher is the original model with the setting from their released codes (\emph{ngf=64})\footnote{``ngf'' indicates ``number of generator filters''}. The student model has the same architecture and depth but fewer channels (\emph{ngf=32/24/16}) compared with its teacher. 
Quantitative experiments have been conducted on Horse $\rightarrow$Zebra, Edge$\rightarrow$Shoe and Cityscapes~\cite{Cordts2016Cityscapes}. Besides, we also conduct qualitative experiments on Winter$\rightarrow$Summer,   Summer$\rightarrow$Winter, Apple$\rightarrow$Orange, Photo$\rightarrow$Monet, Facades and Maps~~\cite{cyclegan_datasets,pix2pix_datasets}.

\vspace{-0.3cm} 

\paragraph{Evaluation Settings}
Following previous works~\cite{teacher_do_more_gankd}, we adopt \emph{Fr\'echet Inception Distance (FID)} and mIoU as the performance metrics on Cityscapes and the other datasets. A lower FID and a higher mIoU indicate that the generated images have better quality. Please refer to the codes in the supplementary material for more details.

\begin{figure*}
    \begin{center}
        \includegraphics[width=17.5cm]{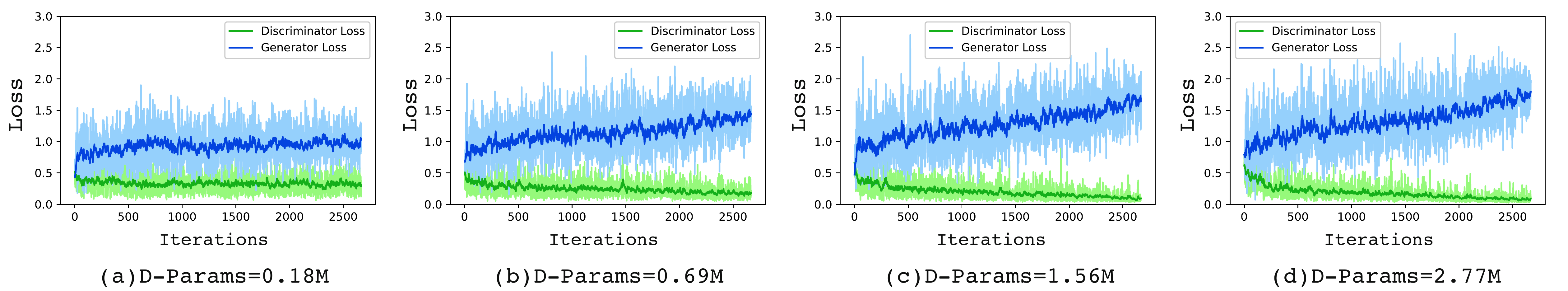}
    \end{center}
    \vspace{-0.5cm}
    \caption{The discriminator loss and generator loss during the training period. In all the subfigures, the generators are 15.81$\times$ compressed. In subfigure (d), the discriminator has its origin size. In subfigure (a-c), the discriminators are compressed by 15.39$\times$, 4.01$\times$ and 1.78$\times$\label{loss}. The FID of these four experiments have been shown in Figure~\ref{fid_diff_d}.
    }
    \vspace{-0.2cm}
\end{figure*}

\begin{figure}
\vspace{-0.25cm}
    \begin{center}
        \includegraphics[width=6.5cm]{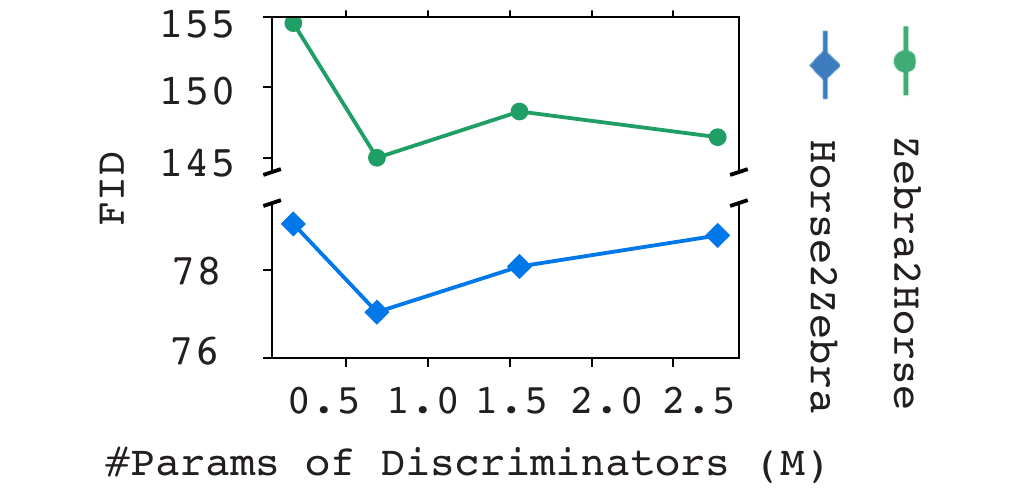}
    \end{center}
     \vspace{-0.5cm}
    \caption{\label{fid_diff_d}Experiments on the distilled CycleGAN with  discriminators in different sizes from Figure~\ref{loss}. Lower FID is better.
     \vspace{-0.3cm}
    }

\end{figure}
\subsection{Quantitative Results}
The quantitative experiment results on paired and unpaired image-to-image translation have been shown in Table~\ref{tab:quan1} and Table~\ref{tab:quan2}, respectively. Experiments on Cityscapes have been shown in Table~\ref{tab:cityscapes}.
It is observed that: \textbf{(a)} Directly applying the native knowledge distillation (Equation~\ref{gan_kd}) to GANs sometimes leads to performance degradation. For instance, there are 1.91 and 0.67 FID increments (performance degradation) on Edges$\rightarrow$Shoes with Pix2pix and Pix2PixHD, respectively.
\textbf{(b)} In contrast, our method achieves consistent and significant performance improvements on all the datasets and models, which outperforms the other GAN knowledge distillation methods by a clear margin, \emph{e.g.} 3.78 FID lower than the second-best method on CycleGAN, on average.
\textbf{(c)} On Horse$\rightarrow$Zebra and Zebra$\rightarrow$Horse, the student models trained with our method achieve almost the same FID with the teacher model, which indicates 7.08$\times$ compression and 6.80$\times$ acceleration with almost no performance degradation.
\textbf{(d)} Compared with the students trained without knowledge distillation, the distilled students usually not only achieve lower FID, but also tend to have lower FID standard deviation, which shows that knowledge distillation may stabilize the training of GANs.
\textbf{(e)} Our method and previous feature knowledge distillation method can be utilized together, which further leads to 0.63 FID reduction on CycleGAN on average.



\subsection{Qualitative Results}
The qualitative results of CycleGAN on Horse$\rightarrow$Zebra (a-d) and Pix2Pix on Edges$\rightarrow$Shoes (e-h) have been shown in Figure~\ref{fig:qualitive}. It is observed that:\textbf{(a)} On Horse$\rightarrow$Zebra, the baseline model can not transform the whole body of horses to zebras (\emph{e.g.} subfigures a, b and c). Besides, the generated stripes of zebras are chaotic and unnatural (\emph{e.g.} subfigure d). This problem also exists in the other knowledge distillation methods (\emph{e.g.} subfigure c).
In contrast, the images generated by the distilled students don't have these issues.
\textbf{(b)} On Edges$\rightarrow$Shoes, the generated images from distilled students have much better color and details (\emph{e.g.} shoestrings in subfigure f and g). In subfigure~(f), the distilled students have successfully generated the highlights on the shoes, which makes the images more realistic.

\vspace{-0.1cm}
\section{Discussion}
\subsection{Ablation Study\label{sec:ablation}}

In this subsection, we give a  detailed study on the individual influence of different frequency bands in knowledge distillation. Experimental results on Horse$\rightarrow $Zebra and Zebra$\rightarrow $Horse with CycleGAN are shown in Table~\ref{tab:ablation}. 
It is observed that: \textbf{(a)} The performance of student models has been severely harmed by only distilling the low frequency band (11.07/10.39 FID increments). \textbf{(b)} The best performance can be achieved by only distilling the high frequency bands (\emph{i.e.} the proposed wavelet knowledge distillation). \textbf{(c)} Distilling both the high and the low frequency bands achieves slight FID reduction, but its performance is still worse than only distilling the high frequency band.
These observations have clearly demonstrated the benefits of distilling the high frequency band and the negative influence of distilling the low frequency band, which also conform to the conclusion in Figure~\ref{fig1} $-$ more attention should be paid to the high frequency band during GAN compression.

\subsection{A Small Discriminator Makes the \\ Compressed Generator Better}
Usually, in real-world GAN applications, only the generators are required to be deployed in devices, while the discriminators are always discarded at this time. As a result, most of the previous works only perform compression on generators but ignore what should be done on discriminators. However, since the discriminator directly influences the training loss of of generators, it 
has a crucial impact on the performance of generators. In this subsection, we
study how the capacity of discriminators influences generators.
Figure~\ref{loss} has shown the training loss of generators and discriminators for four CycleGANs with discriminators in different sizes. In all the subfigures, the generators are $\text{15.81}\times$ compressed and trained with wavelet knowledge distillation.
 In subfigure~(d), the discriminator has its origin size. In subfigure (a-c), the discriminators are compressed by 15.39$\times$, 4.01$\times$, and 1.78$\times$, respectively. Besides, their corresponding FIDs have been shown in Figure~\ref{fid_diff_d}.

\vspace{-0.3cm}
\paragraph{Observation \& Analysis} It is observed that: \textbf{(i)} When the generator is compressed but the discriminator is not compressed (subfigure d), the loss of generator is much higher and the loss of discriminator is much lower. This observation indicates that when the discriminator has a much larger size than the generator, it achieves overwhelming success in its competition with generators. Thus, the balance between discriminators and generators is broken, which makes generators hard to learn useful information from the adversarial loss. \textbf{(ii)} The distilled generator achieves the best performance when the discriminator is 4.01$\times$ compressed (0.69M). Both a too small and a too large discriminator lead to performance degradation on generators,  indicating that the unbalance between discriminators and generators in adversarial learning harms the training of generators. 

Based on these observations, we can conclude that  although discriminators are not utilized in application, they are still required to be properly compressed to maintain the balance between them and  generators in adversarial learning, which further benefits the training of generators. 
\begin{table}[]
    \centering
    \footnotesize
    \caption{Ablation study on different frequency bands. Each experiment is averaged over 8 trials. Lower FID is better.\label{tab:ablation}}
\vspace{-0.2cm}
    \setlength{\tabcolsep}{4.0mm}{\begin{tabular}{cccll}
            \toprule
            \multicolumn{2}{c}{Frequency}&&
            \multicolumn{2}{c}{FID$\downarrow$}\\
            \cmidrule{1-2} \cmidrule{4-5}
            Low&High&&Horse$\rightarrow$Zebra&Zebra$\rightarrow$Horse\\
            \midrule
            \color{red}{$\times$}&$\color{red}{\times}$&&85.04$_{\pm6.88}$&152.67$_{\pm5.07}$\\
            $\color{green}{\checkmark}$&$\color{red}{\times}$&&96.11$_{\pm14.39}$&163.06$_{\pm3.91}$\\
            $\color{red}{\times}$&$\color{green}{\checkmark}$&&\cellcolor{mygray}\textbf{77.04$_{\mathbf{\pm\mathbf{3.52}}}$}&\cellcolor{mygray}\textbf{146.01$_{\mathbf{\pm1.86}}$}\\
            $\color{green}{\checkmark}$&$\color{green}{\checkmark}$&&81.81$_{\pm4.52}$&148.09$_{\pm2.18}$\\
           \bottomrule
          \end{tabular}}
  \vspace{-0.0cm}
\end{table}

\subsection{Knowledge Distillation Paradigm}
\begin{table}[]
\footnotesize
    \caption{Comparison on knowledge distillation paradigms. Wavelet knowledge distillation is utilized in all these experiments.  Each experiment is averaged over 8 trials. Lower FID is better.\label{tab:kd_paradiam}}
    
  \vspace{-0.4cm}
  \begin{center}
      
    \setlength{\tabcolsep}{3.7mm}{\begin{tabular}{cl cc}
            \toprule
            \multirow{2}{*}{Dataset}& \multirow{2}{*}{KD Scheme}&\multicolumn{2}{c}{Metric}\\
            \cmidrule{3-4}
            &&FID$\downarrow$ &$\Delta$\\
                        \midrule
            \multirow{7}{*}{Horse$\rightarrow$Zebra}&Origin Student&85.04$_{\pm6.88}$&--\\
            &\cellcolor{mygray}\textbf{TSKD}&\cellcolor{mygray}\textbf{77.04$_{\pm\mathbf{3.52}}$}&\textbf{8.00}\cellcolor{mygray}\\
            &TAKD$^1$&78.53$_{\pm2.98}$&6.51\\
            &TAKD$^2$&78.69$_{\pm3.26}$&6.35\\
            &SD&83.51$_{\pm2.00}$&1.53\\
            &DML$^1$&81.06$_{\pm3.56}$&3.98\\
            &DML$^2$&84.72$_{\pm5.17}$&0.32\\
            
            \midrule
            \multirow{7}{*}{Zebra$\rightarrow$Horse}&Origin Student&152.67$_{\pm5.07}$&--\\
            &\textbf{TSKD}\cellcolor{mygray}&\cellcolor{mygray}\textbf{146.01$_{\mathbf{\pm1.86}}$}&\cellcolor{mygray}\textbf{6.66}\\
            &TAKD$^1$&148.03$_{\pm1.40}$&4.56\\
            &TAKD$^2$&147.75$_{\pm1.75}$&4.92\\
            &SD&151.74$_{\pm3.46}$&0.93\\
            &DML$^1$&150.37$_{\pm2.18}$&2.30\\
            &DML$^2$&152.03$_{\pm1.93}$&0.64\\
            \bottomrule
          \end{tabular}}
          \end{center}
          {
          \vspace{-0.3cm}
          TAKD$^1$: The FIDs of teacher assistants on Horse $\rightarrow$ Zebra and Zebra $\rightarrow$ Horse are 55.00 and 140.49, respectively.}\\
          {\footnotesize
          TAKD$^2$: The FIDs of teacher assistants on Horse $\rightarrow$ Zebra and Zebra $\rightarrow$ Horse are 51.34 and 133.29, respectively.}\\
          {\footnotesize
          DML$^1$ and DML$^2$: There are 2 and 3 peers
          , respectively.\vspace{-0.2cm}
          }
  
\end{table}
Knowledge distillation is first proposed in a teacher-student {(TSKD)\small} paradigm, where the teacher model is first trained and then distilled to a student model. Recently, abundant knowledge distillation paradigms are proposed to achieve better performance, such as deep mutual learning {\small (DML)}~\cite{deepmutuallearning}, self-distillation {\small (SD)}~\cite{selfdistillation} and teacher-assistant knowledge distillation {\small (TAKD)}~\cite{TAdistillation}. 
Many of these methods lead to higher effectiveness than the traditional {(TSKD)\small} paradigm.
Unfortunately, these {KD\small} paradigms are usually only evaluated on classification tasks and their performance in more challenging tasks has not been well-studied. In this subsection, we have given a comparison of the following {KD\small} paradigms on Image-to-Image translation with {GANs\small}.
{ \small
\begin{itemize}
    \item \emph{TSKD} is the most common KD diagram which trains a large teacher first and then distills it to a small student.
    \item \emph{TAKD} is proposed to bridge the gap between students and teachers with a teacher-assistant. It first distills knowledge from teachers to teacher-assistants and then distills knowledge from teacher assistants to the students~\cite{TAdistillation}. 
    \item \emph{SD} is a special case in TSKD when the student and the teacher have the identical architecture. Experimental and theoretical results have proven its success~\cite{hilbert_kd}.
    \item \emph{DML} (\emph{a.k.a.} online knowledge distillation, collaborative learning) trains several students (\emph{a.k.a.} peers) to learn from each other~\cite{deepmutuallearning}.
\end{itemize}}

\vspace{-0.2cm}
\paragraph{Observation} Experimental results of different knowledge distillation paradigms have been shown in Table~\ref{tab:kd_paradiam}. It is observed that: \textbf{(a)} All the knowledge distillation schemes lead to performance gain compared with the baseline. Besides, the most common {\small TSKD} achieves better performance than the other {\small KD} schemes.
\textbf{(b)} The performance improvements in {\small DML} and {\small SD} are much lower than that in {\small TSKD} and {\small TAKD}, which indicates a pre-trained and high-quality teacher is very crucial on image-to-image translation. \textbf{(c)} There is no significant performance difference between {\small TSKD} and {\small TAKD}, which means that the teacher assistants can not facilitate the training of a tiny student in knowledge distillation on image-to-image translation.

\vspace{-0.5cm}
\paragraph{Analysis} These observations show that there is a huge difference between knowledge distillation on image classification and image-to-image translation. We believe this difference is caused by the following reasons: \textbf{(a)} Compared with image classification, image-to-image translation is more challenging and thus a high performance teacher is more necessary to provide better guidance. \textbf{(b)} Besides, on classification, one of the benefits of the novel knowledge distillation schemes comes from their effectiveness as label smoothing~\cite{2020Revisiting}. However, label smoothing is effective in classification but can not be utilized in image-to-image translation, which is a pixel-level regression problem.

\vspace{-0.1cm}
\section{Conclusion}
\vspace{-0.1cm}
This paper has proposed to analyze and distill GANs on image-to-image translation tasks from a frequency perspective. To the best of our knowledge, we first quantitatively show the difference of GANs performance on different frequency bands and propose to highlight its learning on the high frequency bands during knowledge distillation.  Abundant experiments on both paired and unpaired image-to-image translation have demonstrated its significant performance in terms of both quantitative and qualitative results. For instance, 7.08$\times$ compression and 6.80$\times$ acceleration can be achieved on CycleGAN with almost no performance drop. Experimental results in the ablation study have further shown the merits of distilling the high frequency bands.
Besides, studies on the relation between discriminators and generators in model compression have been introduced, showing that a small discriminator is beneficial during the compression of the generator by maintaining their balance in adversarial learning.
Moreover, we have also analyzed the influence of different knowledge distillation paradigms on GANs for image-to-image translation. Surprisingly, different from the results in classification, most of novel KD paradigms do not work well on GANs. We expect this observation may encourage studies of knowledge distillation in tasks beyond classification. Our limitations and future work are discussed in supplementary materials.

{\small
\bibliographystyle{ieee_fullname}
\bibliography{egbib}

\begin{thebibliography}{10}\itemsep=-1pt

\bibitem{kd_variational}
Sungsoo Ahn, Shell~Xu Hu, Andreas Damianou, Neil~D Lawrence, and Zhenwen Dai.
\newblock Variational information distillation for knowledge transfer.
\newblock In {\em Proceedings of the IEEE Conference on Computer Vision and
  Pattern Recognition}, pages 9163--9171, 2019.

\bibitem{new_style3}
Jie An, Siyu Huang, Yibing Song, Dejing Dou, Wei Liu, and Jiebo Luo.
\newblock Artflow: Unbiased image style transfer via reversible neural flows.
\newblock In {\em {IEEE} Conference on Computer Vision and Pattern Recognition,
  {CVPR} 2021, virtual, June 19-25, 2021}, pages 862--871. Computer Vision
  Foundation / {IEEE}, 2021.

\bibitem{kd_detection4}
Mohammad~Farhadi Bajestani and Yezhou Yang.
\newblock Tkd: Temporal knowledge distillation for active perception.
\newblock In {\em The IEEE Winter Conference on Applications of Computer
  Vision}, pages 953--962, 2020.

\bibitem{bigGAN}
Andrew Brock, Jeff Donahue, and Karen Simonyan.
\newblock Large scale gan training for high fidelity natural image synthesis.
\newblock {\em arXiv preprint arXiv:1809.11096}, 2018.

\bibitem{model_compression}
Cristian Buciluǎ, Rich Caruana, and Alexandru Niculescu-Mizil.
\newblock Model compression.
\newblock In {\em Proceedings of the 12th ACM SIGKDD international conference
  on Knowledge discovery and data mining}, pages 535--541. ACM, 2006.

\bibitem{distill_portable_gan}
Hanting Chen, Yunhe Wang, Han Shu, Changyuan Wen, Chunjing Xu, Boxin Shi, Chao
  Xu, and Chang Xu.
\newblock Distilling portable generative adversarial networks for image
  translation.
\newblock In {\em Proceedings of the AAAI Conference on Artificial
  Intelligence}, volume~34, pages 3585--3592, 2020.

\bibitem{new_style4}
Haibo Chen, Lei Zhao, Zhizhong Wang, Huiming Zhang, Zhiwen Zuo, Ailin Li, Wei
  Xing, and Dongming Lu.
\newblock Dualast: Dual style-learning networks for artistic style transfer.
\newblock In {\em {IEEE} Conference on Computer Vision and Pattern Recognition,
  {CVPR} 2021, virtual, June 19-25, 2021}, pages 872--881. Computer Vision
  Foundation / {IEEE}, 2021.

\bibitem{wae}
Tianshui Chen, Liang Lin, Wangmeng Zuo, Xiaonan Luo, and Lei Zhang.
\newblock Learning a wavelet-like auto-encoder to accelerate deep neural
  networks.
\newblock In {\em Thirty-Second AAAI Conference on Artificial Intelligence},
  2018.

\bibitem{attentiongan}
Xinyuan Chen, Chang Xu, Xiaokang Yang, and Dacheng Tao.
\newblock Attention-gan for object transfiguration in wild images.
\newblock In {\em Proceedings of the European Conference on Computer Vision
  (ECCV)}, pages 164--180, 2018.

\bibitem{new_style2}
Jiaxin Cheng, Ayush Jaiswal, Yue Wu, Pradeep Natarajan, and Prem Natarajan.
\newblock Style-aware normalized loss for improving arbitrary style transfer.
\newblock In {\em {IEEE} Conference on Computer Vision and Pattern Recognition,
  {CVPR} 2021, virtual, June 19-25, 2021}, pages 134--143. Computer Vision
  Foundation / {IEEE}, 2021.

\bibitem{Cordts2016Cityscapes}
Marius Cordts, Mohamed Omran, Sebastian Ramos, Timo Rehfeld, Markus Enzweiler,
  Rodrigo Benenson, Uwe Franke, Stefan Roth, and Bernt Schiele.
\newblock The cityscapes dataset for semantic urban scene understanding.
\newblock In {\em Proc. of the IEEE Conference on Computer Vision and Pattern
  Recognition (CVPR)}, 2016.

\bibitem{wavelet_texture}
Shin Fujieda, Kohei Takayama, and Toshiya Hachisuka.
\newblock Wavelet convolutional neural networks for texture classification.
\newblock {\em arXiv preprint arXiv:1707.07394}, 2017.

\bibitem{gan}
Ian Goodfellow, Jean Pouget-Abadie, Mehdi Mirza, Bing Xu, David Warde-Farley,
  Sherjil Ozair, Aaron Courville, and Yoshua Bengio.
\newblock Generative adversarial nets.
\newblock {\em Advances in neural information processing systems}, 27, 2014.

\bibitem{distill_hinton}
Geoffrey Hinton, Oriol Vinyals, and Jeff Dean.
\newblock Distilling the knowledge in a neural network.
\newblock In {\em NeurIPS}, 2014.

\bibitem{huang2017wavelet}
Huaibo Huang, Ran He, Zhenan Sun, and Tieniu Tan.
\newblock Wavelet-srnet: A wavelet-based cnn for multi-scale face super
  resolution.
\newblock In {\em Proceedings of the IEEE International Conference on Computer
  Vision}, pages 1689--1697, 2017.

\bibitem{pix2pix_datasets}
Phillip Isola, Jun-Yan Zhu, Tinghui Zhou, and Alexei~A Efros.
\newblock Pix2pix datasets,
  \url{http://efrosgans.eecs.berkeley.edu/pix2pix/datasets/}.

\bibitem{pix2pix}
Phillip Isola, Jun-Yan Zhu, Tinghui Zhou, and Alexei~A Efros.
\newblock Image-to-image translation with conditional adversarial networks.
\newblock In {\em Proceedings of the IEEE conference on computer vision and
  pattern recognition}, pages 1125--1134, 2017.

\bibitem{DBLP:conf/cvpr/JeongKLS21}
Somi Jeong, Youngjung Kim, Eungbean Lee, and Kwanghoon Sohn.
\newblock Memory-guided unsupervised image-to-image translation.
\newblock In {\em {IEEE} Conference on Computer Vision and Pattern Recognition,
  {CVPR} 2021, virtual, June 19-25, 2021}, pages 6558--6567. Computer Vision
  Foundation / {IEEE}, 2021.

\bibitem{teacher_do_more_gankd}
Qing Jin, Jian Ren, Oliver~J Woodford, Jiazhuo Wang, Geng Yuan, Yanzhi Wang,
  and Sergey Tulyakov.
\newblock Teachers do more than teach: Compressing image-to-image models.
\newblock In {\em Proceedings of the IEEE/CVF Conference on Computer Vision and
  Pattern Recognition}, pages 13600--13611, 2021.

\bibitem{stylegan}
Tero Karras, Samuli Laine, and Timo Aila.
\newblock A style-based generator architecture for generative adversarial
  networks.
\newblock In {\em Proceedings of the IEEE/CVF Conference on Computer Vision and
  Pattern Recognition}, pages 4401--4410, 2019.

\bibitem{styleganv2}
Tero Karras, Samuli Laine, Miika Aittala, Janne Hellsten, Jaakko Lehtinen, and
  Timo Aila.
\newblock Analyzing and improving the image quality of stylegan.
\newblock In {\em Proceedings of the IEEE/CVF Conference on Computer Vision and
  Pattern Recognition}, pages 8110--8119, 2020.

\bibitem{srgan}
Christian Ledig, Lucas Theis, Ferenc Husz{\'a}r, Jose Caballero, Andrew
  Cunningham, Alejandro Acosta, Andrew Aitken, Alykhan Tejani, Johannes Totz,
  Zehan Wang, et~al.
\newblock Photo-realistic single image super-resolution using a generative
  adversarial network.
\newblock In {\em Proceedings of the IEEE conference on computer vision and
  pattern recognition}, pages 4681--4690, 2017.

\bibitem{gan_compress}
Muyang Li, Ji Lin, Yaoyao Ding, Zhijian Liu, Jun-Yan Zhu, and Song Han.
\newblock Gan compression: Efficient architectures for interactive conditional
  gans.
\newblock In {\em Proceedings of the IEEE/CVF Conference on Computer Vision and
  Pattern Recognition}, pages 5284--5294, 2020.

\bibitem{revisit_discriminator}
Shaojie Li, Jie Wu, Xuefeng Xiao, Fei Chao, Xudong Mao, and Rongrong Ji.
\newblock Revisiting discriminator in {GAN} compression: {A}
  generator-discriminator cooperative compression scheme.
\newblock {\em CoRR}, abs/2110.14439, 2021.

\bibitem{DBLP:conf/cvpr/LiZ0CHMHWJ21}
Xinyang Li, Shengchuan Zhang, Jie Hu, Liujuan Cao, Xiaopeng Hong, Xudong Mao,
  Feiyue Huang, Yongjian Wu, and Rongrong Ji.
\newblock Image-to-image translation via hierarchical style disentanglement.
\newblock In {\em {IEEE} Conference on Computer Vision and Pattern Recognition,
  {CVPR} 2021, virtual, June 19-25, 2021}, pages 8639--8648. Computer Vision
  Foundation / {IEEE}, 2021.

\bibitem{spkd_gan}
Zeqi Li, Ruowei Jiang, and Parham Aarabi.
\newblock Semantic relation preserving knowledge distillation for
  image-to-image translation.
\newblock In {\em European Conference on Computer Vision}, pages 648--663.
  Springer, 2020.

\bibitem{lin2020weight}
Ye Lin, Yanyang Li, Ziyang Wang, Bei Li, Quan Du, Tong Xiao, and Jingbo Zhu.
\newblock Weight distillation: Transferring the knowledge in neural network
  parameters.
\newblock {\em arXiv preprint arXiv:2009.09152}, 2020.

\bibitem{liu2019paying}
Miao Liu, Xin Chen, Yun Zhang, Yin Li, and James~M Rehg.
\newblock Attention distillation for learning video representations.
\newblock In {\em BMVC}, 2020.

\bibitem{liu2018multi}
Pengju Liu, Hongzhi Zhang, Kai Zhang, Liang Lin, and Wangmeng Zuo.
\newblock Multi-level wavelet-cnn for image restoration.
\newblock In {\em Proceedings of the IEEE conference on computer vision and
  pattern recognition workshops}, pages 773--782, 2018.

\bibitem{new_style5}
Xiao{-}Chang Liu, Yong{-}Liang Yang, and Peter Hall.
\newblock Learning to warp for style transfer.
\newblock In {\em {IEEE} Conference on Computer Vision and Pattern Recognition,
  {CVPR} 2021, virtual, June 19-25, 2021}, pages 3702--3711. Computer Vision
  Foundation / {IEEE}, 2021.

\bibitem{structured_kd}
Yifan Liu, Ke Chen, Chris Liu, Zengchang Qin, Zhenbo Luo, and Jingdong Wang.
\newblock Structured knowledge distillation for semantic segmentation.
\newblock In {\em Proceedings of the IEEE Conference on Computer Vision and
  Pattern Recognition}, pages 2604--2613, 2019.

\bibitem{liu2021content}
Yuchen Liu, Zhixin Shu, Yijun Li, Zhe Lin, Federico Perazzi, and Sun-Yuan Kung.
\newblock Content-aware gan compression.
\newblock In {\em Proceedings of the IEEE/CVF Conference on Computer Vision and
  Pattern Recognition}, pages 12156--12166, 2021.

\bibitem{wavelet}
St{\'e}phane Mallat.
\newblock {\em A wavelet tour of signal processing}.
\newblock Elsevier, 1999.

\bibitem{TAdistillation}
Seyed-Iman Mirzadeh, Mehrdad Farajtabar, Ang Li, and Hassan Ghasemzadeh.
\newblock Improved knowledge distillation via teacher assistant: Bridging the
  gap between student and teacher.
\newblock {\em arXiv preprint arXiv:1902.03393}, 2019.

\bibitem{hilbert_kd}
Hossein Mobahi, Mehrdad Farajtabar, and Peter~L Bartlett.
\newblock Self-distillation amplifies regularization in hilbert space.
\newblock {\em arXiv preprint arXiv:2002.05715}, 2020.

\bibitem{breakingcycle}
Ori Nizan and Ayellet Tal.
\newblock Breaking the cycle-colleagues are all you need.
\newblock In {\em Proceedings of the IEEE/CVF Conference on Computer Vision and
  Pattern Recognition}, pages 7860--7869, 2020.

\bibitem{distillationdefense}
Nicolas Papernot, Patrick McDaniel, Xi Wu, Somesh Jha, and Ananthram Swami.
\newblock Distillation as a defense to adversarial perturbations against deep
  neural networks.
\newblock In {\em IEEE Symposium on Security and Privacy}, pages 582--597,
  2016.

\bibitem{park2020contrastive}
Taesung Park, Alexei~A Efros, Richard Zhang, and Jun-Yan Zhu.
\newblock Contrastive learning for unpaired image-to-image translation.
\newblock In {\em European Conference on Computer Vision}, pages 319--345.
  Springer, 2020.

\bibitem{gaugan}
Taesung Park, Ming-Yu Liu, Ting-Chun Wang, and Jun-Yan Zhu.
\newblock Gaugan: semantic image synthesis with spatially adaptive
  normalization.
\newblock In {\em ACM SIGGRAPH 2019 Real-Time Live!}, pages 1--1. 2019.

\bibitem{relational_kd}
Wonpyo Park, Dongju Kim, Yan Lu, and Minsu Cho.
\newblock Relational knowledge distillation.
\newblock In {\em Proceedings of the IEEE Conference on Computer Vision and
  Pattern Recognition}, pages 3967--3976, 2019.

\bibitem{DBLP:conf/cvpr/RichardsonAPNAS21}
Elad Richardson, Yuval Alaluf, Or Patashnik, Yotam Nitzan, Yaniv Azar, Stav
  Shapiro, and Daniel Cohen{-}Or.
\newblock Encoding in style: {A} stylegan encoder for image-to-image
  translation.
\newblock In {\em {IEEE} Conference on Computer Vision and Pattern Recognition,
  {CVPR} 2021, virtual, June 19-25, 2021}, pages 2287--2296. Computer Vision
  Foundation / {IEEE}, 2021.

\bibitem{kd_bert1}
Victor Sanh, Lysandre Debut, Julien Chaumond, and Thomas Wolf.
\newblock Distilbert, a distilled version of bert: smaller, faster, cheaper and
  lighter.
\newblock {\em arXiv preprint arXiv:1910.01108}, 2019.

\bibitem{singan}
Tamar~Rott Shaham, Tali Dekel, and Tomer Michaeli.
\newblock Singan: Learning a generative model from a single natural image.
\newblock In {\em Proceedings of the IEEE/CVF International Conference on
  Computer Vision}, pages 4570--4580, 2019.

\bibitem{kd_crd}
Yonglong Tian, Dilip Krishnan, and Phillip Isola.
\newblock Contrastive representation distillation.
\newblock {\em arXiv preprint arXiv:1910.10699}, 2019.

\bibitem{relational_kd2}
Frederick Tung and Greg Mori.
\newblock Similarity-preserving knowledge distillation.
\newblock In {\em Proceedings of the IEEE International Conference on Computer
  Vision}, pages 1365--1374, 2019.

\bibitem{videogan}
Carl Vondrick, Hamed Pirsiavash, and Antonio Torralba.
\newblock Generating videos with scene dynamics.
\newblock {\em Advances in neural information processing systems}, 29:613--621,
  2016.

\bibitem{new_style1}
Pei Wang, Yijun Li, and Nuno Vasconcelos.
\newblock Rethinking and improving the robustness of image style transfer.
\newblock In {\em {IEEE} Conference on Computer Vision and Pattern Recognition,
  {CVPR} 2021, virtual, June 19-25, 2021}, pages 124--133. Computer Vision
  Foundation / {IEEE}, 2021.

\bibitem{pix2pixHD}
Ting-Chun Wang, Ming-Yu Liu, Jun-Yan Zhu, Andrew Tao, Jan Kautz, and Bryan
  Catanzaro.
\newblock High-resolution image synthesis and semantic manipulation with
  conditional gans.
\newblock In {\em Proceedings of the IEEE conference on computer vision and
  pattern recognition}, pages 8798--8807, 2018.

\bibitem{wavepool}
Travis Williams and Robert Li.
\newblock Wavelet pooling for convolutional neural networks.
\newblock In {\em International Conference on Learning Representations}, 2018.

\bibitem{kd_bert2}
Canwen Xu, Wangchunshu Zhou, Tao Ge, Furu Wei, and Ming Zhou.
\newblock Bert-of-theseus: Compressing bert by progressive module replacing.
\newblock {\em arXiv preprint arXiv:2002.02925}, 2020.

\bibitem{2020Revisiting}
L. Yuan, F.~E. Tay, G. Li, T. Wang, and J. Feng.
\newblock Revisiting knowledge distillation via label smoothing regularization.
\newblock In {\em 2020 IEEE/CVF Conference on Computer Vision and Pattern
  Recognition (CVPR)}, 2020.

\bibitem{attentiondistillation}
Sergey Zagoruyko and Nikos Komodakis.
\newblock Paying more attention to attention: Improving the performance of
  convolutional neural networks via attention transfer.
\newblock In {\em ICLR}, 2017.

\bibitem{detectiondistillation}
Linfeng Zhang and Ma Kaisheng.
\newblock Improve object detection with feature-based knowledge distillation:
  Towards accurate and efficient detectors.
\newblock In {\em ICLR}, 2021.

\bibitem{tofd}
Linfeng Zhang, Yukang Shi, Zuoqiang Shi, Kaisheng Ma, and Chenglong Bao.
\newblock Task-oriented feature distillation.
\newblock In {\em NeurIPS}, 2020.

\bibitem{selfdistillation}
Linfeng Zhang, Jiebo Song, Anni Gao, Jingwei Chen, Chenglong Bao, and Kaisheng
  Ma.
\newblock Be your own teacher: Improve the performance of convolutional neural
  networks via self distillation.
\newblock In {\em arXiv preprint:1905.08094}, 2019.

\bibitem{Zhang2019SCANAS}
Linfeng Zhang, Zhanhong Tan, Jiebo Song, Jingwei Chen, Chenglong Bao, and
  Kaisheng Ma.
\newblock Scan: A scalable neural networks framework towards compact and
  efficient models.
\newblock {\em ArXiv}, abs/1906.03951, 2019.

\bibitem{deepmutuallearning}
Ying Zhang, Tao Xiang, Timothy~M Hospedales, and Huchuan Lu.
\newblock Deep mutual learning.
\newblock In {\em CVPR}, pages 4320--4328, 2018.

\bibitem{cyclegan_datasets}
Jun-Yan Zhu, Taesung Park, Phillip Isola, and Alexei~A Efros.
\newblock Pix2pix datasets,
  \url{http://efrosgans.eecs.berkeley.edu/cyclegan/datasets}.

\bibitem{cyclegan}
Jun-Yan Zhu, Taesung Park, Phillip Isola, and Alexei~A Efros.
\newblock Unpaired image-to-image translation using cycle-consistent
  adversarial networks.
\newblock In {\em Proceedings of the IEEE international conference on computer
  vision}, pages 2223--2232, 2017.

\end{thebibliography}
}

\end{document}